\documentclass[letterpaper, 10 pt, conference]{ieeeconf}  

\IEEEoverridecommandlockouts                              

\usepackage{amsmath,amssymb,graphicx, multirow}
\usepackage{flushend}
\usepackage{float}
\usepackage{mathtools}
 \usepackage{url}
\usepackage{color}

\title{\LARGE \bf
Boosting Image Super-Resolution Via Fusion of Complementary Information Captured by Multi-Modal Sensors}

\author{Fan Wang$^{1, 2}$, Jiangxin Yang$^{1, 2}$, Yanlong Cao$^{1, 2}$, Yanpeng Cao$^{1, 2,*}$, and Michael Ying Yang$^{3}$
\thanks{$^{1}$State Key Laboratory of Fluid Power and Mechatronic Systems, School of  Mechanical Engineering, Zhejiang University, Hangzhou, China}%
\thanks{$^{2}$Key Laboratory of Advanced Manufacturing Technology of Zhejiang Province, School of Mechanical Engineering,  Zhejiang University, Hangzhou, China}%
\thanks{$^{3}$Scene Understanding Group, University of Twente, Hengelosestraat 99, 7514 AE Enschede, The Netherlands}
\thanks{$^{*}$Corresponding author}%
}

\begin{document}
\maketitle
\thispagestyle{empty}
\pagestyle{empty}

\begin{abstract}


Image Super-Resolution (SR) provides a promising technique to enhance the image quality of low-resolution optical sensors, facilitating better-performing target detection and autonomous navigation in a wide range of robotics applications. It is noted that the state-of-the-art SR methods are typically trained and tested using single-channel inputs, neglecting the fact that the cost of capturing high-resolution images in different spectral domains varies significantly. In this paper, we attempt to leverage complementary information from a low-cost channel (visible/depth) to boost image quality of an expensive channel (thermal) using fewer parameters. To this end, we first present an effective method to virtually generate pixel-wise aligned visible and thermal images based on real-time 3D reconstruction of multi-modal data captured at various viewpoints. Then, we design a feature-level multispectral fusion residual network model to perform high-accuracy SR of thermal images by adaptively integrating co-occurrence features presented in multispectral images. Experimental results demonstrate that this new approach can effectively alleviate the ill-posed inverse problem of image SR by taking into account complementary information from an additional low-cost channel, significantly outperforming state-of-the-art SR approaches in terms of both accuracy and efficiency.
\end{abstract}

\begin{figure*}[ht]
	\centering
	\centerline{\includegraphics[width=0.8\linewidth]{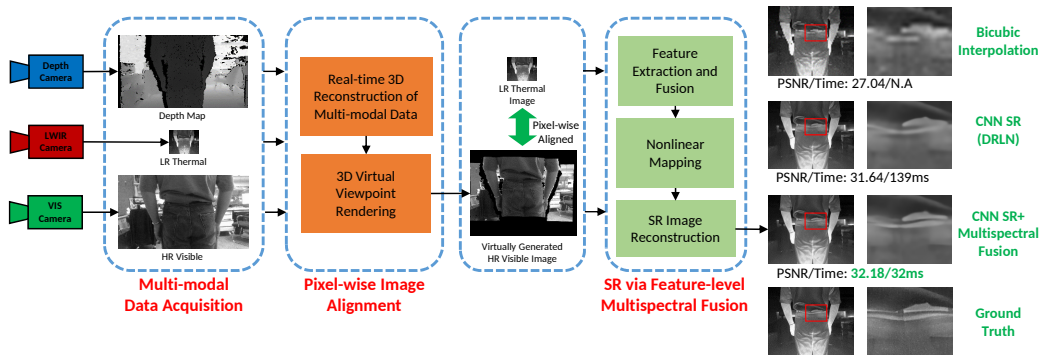}}
	\caption{Overview of our proposed method to boost the accuracy of SR of the high-cost thermal images by taking into account the complementary information captured by an additional low-cost visible sensor. Compared with the state-of-the-art SR methods (e.g., DRLN\cite{anwar2020densely}), our proposed method can achieve higher SR accuracy (PSNR: 32.18dB vs. 31.64dB) and faster running time (32ms vs. 139ms).}
	\label{fig:Compair1}
\end{figure*}

\section{INTRODUCTION}

In recent years, image super-resolution (SR) has attracted considerable attention from both the academic and industrial communities, increasing the spatial resolution of optical sensors. A substantial number of SR methods have been proposed to reconstruct a high-resolution (HR) image based on its low-resolution (LR) version such as self-exemplars method \cite{SelfExSR}, anchored neighborhood regression methods \cite{Timofte2013Anchored}, sparse representation \cite{Yang2012Coupled}, random forest \cite{schulter2015fast}. The restored HR images with high-fidelity details lead to performance gains of various robot vision applications such as target detection, object recognition, medical diagnosis, and autonomous driving \cite{huang2017simultaneous, haut2018new, pillai2019superdepth, islam2020underwater}. 

Due to the recent successes of deep learning in various applications (e.g., speech recognition, image classification, and scene perception), convolutional neural networks (CNNs) have been adopted to solve the ill-posed SR problem \cite{SRCNN2014ECCV, Ledig2017SRGAN, EDSR, zhang2018image}. Dong et al. present a three-layer SRCNN model to learn a nonlinear LR-to-HR mapping function, which is the first SR method adopting CNN architecture and significantly outperforms the classical machine learning-based methods \cite{SRCNN2014ECCV}. Later extensions of SRCNN mostly improve SR accuracy by deploying more complex network architectures (e.g., D-DBPN\cite{Haris2018}, RCAN\cite{zhang2018image}, or DRLN \cite{anwar2020densely}) or/and using a higher quality training dataset (e.g., DIV2K \cite{NTIRE2017} or ImageNet \cite{ILSVRC15}). However, the training of a deeper network is a difficult task due to the gradient vanishing/exploring problem. Moreover, the recently proposed CNN-based SR models typically contain a large number of parameters which are impractical for hardware implementation and real-time execution \cite{zhang2018image, anwar2020densely, dai2019second}. Therefore, the motivation of this paper is to explore other alternatives to achieve significant improvement for the SR task without reproducing some proven effective techniques such as increasing network depth. 

It is noted that the existing SR methods are typically trained/tested using single-channel inputs, neglecting the fact that the cost of capturing HR images in various spectral domains (e.g., visible, short-wave infrared, long-wave infrared) are significantly different. For instance, long-wave infrared (LWIR) detectors are encapsulated in individual vacuum packages to achieve high-accuracy thermal measurement which is a time-consuming and expensive process \cite{rogalski2016challenges}. As the result, LWIR sensors are significantly more expensive than the visible ones with a similar resolution. Commercial LWIR cameras typically capture LR images (e.g., $160\times120$ or even $80\times60$ pixels), in which important thermal details are lost and some target objects become difficult to detect/recognize. 

In this paper, we attempt to boost the performance of image restoration in the expensive thermal channel by taking into account the complementary information captured by an additional low-cost visible sensor. The major challenge is that visible and thermal images are captured using two individual cameras and in very different spectrum ranges (visible and LWIR sensors working in 380-780nm and 8-14$\mu$m, respectively). Therefore, it is practically difficult to obtain pixel-wise registered visible and thermals by adjusting the poses of two individual cameras. Also, the captured multispectral images present very different characteristics thus it is difficult to accurately restore fine thermal details without adding irrelevant features extracted in the visible images. To overcome the aforementioned problems, we first present an effective method to perform fast and accurate 3D registration of multi-modal data captured at various viewpoints and generate pixel-wise aligned visible and thermal images via virtual viewpoint rendering. Also, we propose a feature-level multispectral fusion residual network (FL-MFRN) architecture to achieve high-accuracy SR results of thermal images by adaptively integrating co-occurrence features (e.g., structural edges and texture details) presented in multispectral images.  As illustrated in Fig.~\ref{fig:Compair1}, important thermal details, which are difficult/impossible to restore based on the LR thermal image alone, can be accurately reconstructed by utilizing texture/edge features in the visible channel. The contributions of this paper are mainly summarized as follows:

First, we design a feature-level multispectral fusion CNN model to perform high-quality SR. To the best of our knowledge, it represents the first research work successfully demonstrating the feature-level fusion of images captured in very different spectrum ranges (380-780nm visible and 8-14$\mu$m LWIR) leads to better SR results in terms of higher accuracy, better visual perception, and less computational load. 

Second, we build a multi-modal imaging system consisting of visible, LWIR, and depth sensors and present an effective method to perform fast and accurate 3D registration of multi-modal data. Visible and thermal images captured at similar viewpoints are associated to generate pixel-wise aligned image pairs in complex 3D scenes via virtual viewpoint rendering.

Third, our method achieves significantly higher image restoration accuracy compared with the state-of-the-art SR methods \cite{chen2016color, dai2019second, anwar2020densely, zhang2018image}. Moreover, this efficient model can process ~30 image pairs per second on a single NVIDIA GeForce RTX 2080Ti GPU to facilitate real-time SR applications. Codes and the captured multi-modal dataset will be made available.


\section{RELATED WORKS}
\label{related}

Recently, deep CNN-based models have been successfully used to achieve breakthrough improvements in various computer vision and robotics applications\cite{Weaklysupervised2017CVPR, lin2016IJCV, Deeppose2014CVPR, noh2015ICCV}. Many researchers also attempt to solve the SR problem through the training of the LR to HR  mapping functions using large training datasets. Dong et al. proposed the first CNN-based SR model (SRCNN) \cite{SRCNN2014ECCV}, which is based on a compact three-layer lightweight structure but significantly outperforms other machine learning-based methods \cite{Yang2008Image, Yang2012Coupled, Timofte2013Anchored, Bevilacqua2012}. Following this pioneering work, many researchers attempted to achieve more accurate SISR results by either increasing the depth of the network or deploying more complex architectures. State-of-the-art SR performances are achieved based on very deep CNN models such as RDN \cite{RDN}, D-DBPN\cite{Haris2018}, RCAN\cite{zhang2018image}, DRLN \cite{anwar2020densely}, and SAN \cite{dai2019second} which are heavily trained using the high-resolution DIV2K \cite{NTIRE2017} dataset (containing 800 training images of 2K resolution) or ImageNet \cite{ILSVRC15} subset. However, their training processes take a long time to complete as well as the predicting processes. A noticeable drawback of the above-mentioned SR methods is that they contain a large number of network parameters and require heavy computational loads.


Due to the complicated fabrication process, LWIR sensors are significantly more expensive than visible ones with similar spatial resolution. Although many successful SR methods have been proposed to increase the resolution of visible images, the development of SR models working well for LR thermal images is surprisingly under-explored. Since visible and thermal images present very different visual characteristics, SR models trained in the visible domain cannot achieve satisfactory reconstruction results when applied to thermal images \cite{Ten}. He et al. constructed an thermal image dataset (120 thermal images of $640\times480$ pixels) which can be used to learn the complex mapping relationship between LR and HR thermal images \cite{CDN-MRF}. Moreover, many deep-learning-based SR solutions \cite{VDSR, SRCNN2015PAMI, Ten, SRCNN2014ECCV, Perceptual-losses} only demonstrate satisfactory reconstruction results for small scale factor SR tasks ($\times2$ or $\times4$) which might not be sufficient for LR thermal images (e.g., $80 \times 60$). Lai et al. proposed a Laplacian Pyramid Super-Resolution Network (LapSRN) to reconstruct the high-frequency residuals at multiple scales progressively, generating high-quality $\times8$ visible SR results \cite{LapSRN}. He et al. presented a cascaded architecture, setting up a mid-point (scale $\times2$) between scale $\times1$ and $\times8$, to accurately rebuild $\times8$ thermal images \cite{CDN-MRF}. However, large scale factor SR still remains a challenging problem since the original information in the HR image has little or no evidence in its LR version \cite{Perceptual-losses}.

In recent years, several multispectral/hyperspectral fusion solutions have been reported in the literature \cite{lanaras2018super,yang2017pannet,lanaras2017hyperspectral}. It is noted that many of these methods aim to generate an informative fused output by integrating the most distinctive features extracted in multiple channels \cite{lanaras2017hyperspectral}. In this paper, we focus on the high-accuracy SR of thermal images, which is particularly important for many thermal monitoring/diagnosis applications. It worth mentioning that some CNN-based pan-sharpening algorithms have recently been proposed for up-scaling LR hyperspectral images using a corresponding HR panchromatic (PAN) image (i.e., grayscale) \cite{yang2017pannet, hu2020pan}. However, these methods typically employ a pixel-level fusion architecture and only perform well on images captured in similar or overlapping spectrum ranges (e.g., 380-780nm visible and 780-1000nm NIR). To the best of our knowledge, this paper represents the first research work successfully demonstrating the feature-level fusion of images captured in very different spectrum ranges (380-780nm visible and 8-14$\mu$m LWIR) leads to better SR results in terms of higher accuracy, better visual perception, and faster running time. 

\section{Pixel-wise Alignment of Multispectral Images}
\label{method1}

Visible and thermal images are captured using two individual cameras working in very different spectrum ranges. In setups of beam-splitter or binocular-vision \cite{Ha2017MFNet, Hwang2015, gonzalez2016pedestrian}, it is practically difficult to obtain pixel-wise aligned visible and thermal images at various depths by adjusting the poses of two individual cameras. To address this issue, we present an effective method to generate pixel-wise aligned visible and thermal images based on 3D registration of multi-modal data and virtual viewpoint rendering.

\begin{figure}
	\centering
	\centerline{\includegraphics[width=0.8\linewidth]{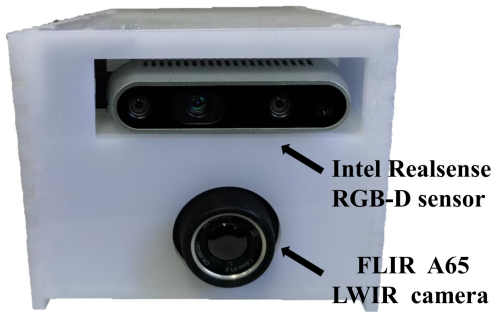}}
	\caption{FLIR A65 LWIR and Intel Realsense RGB-D cameras are rigidly attached using an acrylic frame for multi-modal data capturing.}
	\label{KinectXenics}
\end{figure} 

We set up a multi-modal imaging system consisting of a FLIR A65 LWIR camera (working spectral band is 8-14 $\mu m$) and an Intel Realsense RGB-D sensor. Two sensors are rigidly attached to preserve their relative position and orientation as shown in Fig.~\ref{KinectXenics}. The resolution of the LWIR camera is $640\times512$ pixels and its frame rate is 30 fps. The RGB-D camera simultaneously captures a $1280\times720$ depth image and a $1280\times720$ color image, both at a rate of 30 fps. We keep the original $640\times512$ thermal image and remove the border area of the visible image to make two images have the same field of view. The cropped visible images are resized to $640\times512$ resolution through the bicubic interpolation. We made a target board made of materials with different emissivities \cite{nakagawa2014visualization} and then applied the algorithm presented by Zhang et al. \cite{zhang2000flexible} to compute intrinsic matrices of visible, LWIR, and depth cameras. Finally, we adopted the calibration technique described in \cite{vidas2012mask} to compute extrinsic matrices among multi-modal sensors. 

We applied the Thermal-guided ICP method \cite{cao2018depth} to perform real-time 3D reconstruction of multi-modal data captured at various viewpoints and compute accurate trajectories of visible, LWIR, and depth cameras. This method employs an effective coarse-to-fine methodology to improve the robustness of camera pose estimation which enables to handle significant camera motion. Firstly, the initial pose of the sensing device is estimated using correspondences identified through maximizing the thermal consistency of two consecutive thermal images ${I}^{t-1}_T$ and ${I}^{t}_T$ as
\begin{equation}
{E}_{tc} ( \mathbf{u}) =\sum_{\mathbf{x} \in {\Omega}} \left ( {I}^t_T(\mathbf{x} + \mathbf{u}) - {I}^{t-1}_T(\mathbf{x}) \right ) ^ 2,
\label{myeq1}
\end{equation}
where $\mathbf{x} $ denotes pixel coordinates on the 2D thermal image plane ${\Omega}$ and $\mathbf{u}$ is a 2D displacement vector maximizing the thermal consistency between ${I}^{t-1}_T$ and ${I}^{t}_T$. We use the computed optimal 2D translation $\mathbf{\widehat{u}}$ to robustly establish correspondences between two wide baseline 3D models. Then, the coarse estimation results are further refined by finding the motion parameters that minimize a combined geometric and thermographic loss function. The motion parameters are found by maximizing both geometric and thermographic consistencies between sequential thermal-depth data. Two cost functions, which correspond to geometric and thermographic information respectively, are combined in a weighted sum as
\begin{equation}
	{E}(\mathbf{T}_{D}) = {E}_{icp}(\mathbf{T}_{D}) + \omega {E}_{td}(\mathbf{T}_{D}),
\label{myeq4}
\end{equation}
where $\mathbf{T}_{D}$ is the estimated global pose of the depth camera, $\omega$ is the weight which is set empirically to 0.1 to balance the strength of the ICP term ${E}_{icp}$ and thermal term ${E}_{td}$. The 3D trajectories of visible and LWIR cameras are also computed by referring to their extrinsic matrices. By utilizing complementary information captured by multimodal sensors, our proposed method can handle large camera motion and generate more accurate 3D reconstruction results. 

\begin{figure}[ht]
	\footnotesize
	\centering
	\centerline{\includegraphics[width=0.99\linewidth]{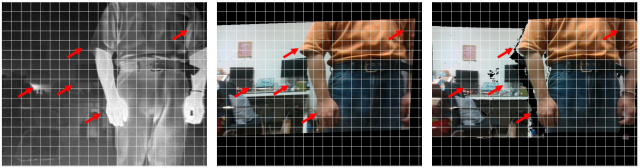}}
    \caption{An illustration of pixel-wise alignment between visible and thermal images. Left: Thermal image; Middle: Alignment result based on homography matrix warping; Right: Alignment result of our proposed method. Please zoom in to check regions highlighted using red arrows.}
	\label{3D_alignment}
\end{figure}

For a thermal image captured at a specific viewpoint $\mathbf{T}_{T}$, we identify the visible image captured at the most similar position ($\mathbf{T}_{V}\approx\mathbf{T}_{T}$) based on the computed 3D trajectories of visible and LWIR cameras to maximize their FOV overlap. For a pixel in the visible image $\mathbf{x}_V(i)$, its corresponding 3D volume element (voxel) $\mathbf{P}(i)$ in the visual hull is efficiently identified using the ray casting algorithm \cite{matusik2001polyhedral}. Then, the voxel $\mathbf{P}(i)$ is transformed to the coordinate system of the LWIR camera and projected to the thermal image plane as
\begin{equation}
\mathbf{x}_T(i)=\mathbf{\kappa}_T(\mathbf{T}_T^{-1} \mathbf{T}_V \mathbf{P}(i)),
\end{equation}
where $\mathbf{x}_T(i)$ is the computed pixel coordinate on the 2D thermal image, and $\mathbf{\kappa}_T$ is the function encoding intrinsic parameters of the LWIR camera for 3D-to-2D projection. As illustrated in Fig.~\ref{3D_alignment}, we can make use of the computed $\mathbf{x}_T$ to generate a pixel-wise aligned visible image for the thermal one via virtual viewpoint rendering. Note our proposed method is fully automatic and does not require manual selection of corresponding points to calculate a homography matrix for image warping.

\section{Feature-level Multispectral Fusion for SR}
\label{method2}

\begin{figure*}[ht]
	\centering
	\includegraphics[width=0.95\linewidth]{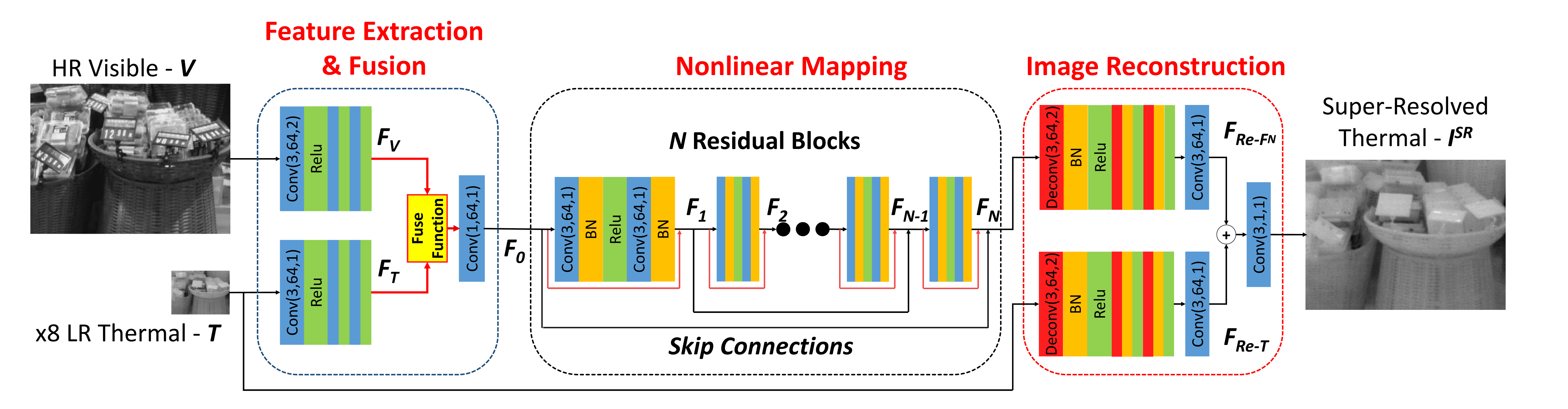}
    \caption{The architecture of our proposed FL-MFRN for SR which contains three parts: feature extraction, nonlinear mapping, and image reconstruction. $Conv(k,n,s)$ indicates that the convolution uses $n$ kernels of size $k\times k$ on the images/features with stride $s$.}
	\label{fig:workflow}
\end{figure*}

Based on the most commonly used CNN-based architecture for SR (Residual Network \cite{EDSR, Resnet, DRRN, anwar2020densely}), we design a feature-level multispectral fusion residual network (FL-MFRN) architecture to achieve high-accuracy SR of thermal images. As shown in Fig.~\ref{fig:workflow}, a HR visible image $V$ (here we only consider the luminance channel - Y channel in YCbCr color space) and a $\times8$ scale LR thermal image $T$ are passed into the feature extraction module to extract low-level features in each stream. Each feature extraction stream consists of three convolutional layers and another three rectified linear unit (ReLU) layers are added after convolution operation to embed more nonlinear terms into the network. It is noted that convolution employs $64$ kernels of size $3\times 3$ on the images/features using stride $1$ in the thermal stream while using stride $2$ in the visible stream. As the result, HR visible images are down-scaled to $1/8$ of its original size to decrease the computational complexity.

We deploy a fusion layer (Fuse function) after the two-stream feature extraction module to combine the low-level feature maps extracted on the visible and thermal channels. The multispectral feature maps are obtained as
\begin{equation}
\mathbf{F}=f(\mathbf{F_V},\mathbf{F_T}),
\label{eq5}
\end{equation}
where $f$ is Summation function to add multispectral feature maps, $\mathbf{F_V}$ and $\mathbf{F_T}$ are the feature maps extracted on the visible and thermal images, and $\mathbf{F}$ is the fused multispectral feature. Finally, a $1\times1$ convolutional layer is adopted to adjust channel weights for the initially combined multispectral feature maps. To process images captured in very different spectrum ranges (380-780nm visible and 8-14$\mu$m LWIR), we propose to extract features in the visible and thermal channels individually for feature-level fusion instead of performing pixel-level fusion and single-stream multispectral feature extraction.



The fused low-level multispectral feature map $\mathbf{F}_0$ is then fed into a number of stacked residual blocks to extract high-level features for HR thermal image reconstruction. Without loss of generality, we employ the residual block used in SRResNet \cite{Ledig2017SRGAN} which consists of two convolutional layers, two batch normalization (BN) layers, one ReLU, and one element-wise addition. The $i_{th}$ residual block takes feature maps $F_{i-1}$ as the input and computes
\begin{equation}
\mathbf{F}_{i} = Res(\mathbf{F}_{i-1}), i=\{1,2,\cdots,N\},
\end{equation}
where $Res(\cdot)$ indicates operations within a single residual block. Inspired by \cite{Mao2016NIPS, DRCN}, we add a number of skip connections between shallower and deeper layers as illustrated in Fig.~\ref{fig:workflow}. Adding skip connections between multiple-stacked layers enables gradient signal to back-propagate directly from the higher-level features to lower-level ones, alleviating the gradient vanishing/exploring problem\cite{Mao2016NIPS}.

After adding $N$ residual blocks, our FL-MFRN model also embeds a number of deconvolution layers at the end of the network to reconstruct the final HR output $I_{SR}$. In a cascaded manner, three individual deconvolution layers are employed to up-scale the semantic multispectral feature maps $F_N $ and the LR thermal image $T$ from $\times 1$ to $\times 2$, $\times 2$ to $\times 4$ and $\times 4$ to $\times 8$, respectively. The advantage of deploying a deconvolution layer is two-fold. First, it avoids artifacts induced by hand-crafted image pre-processing techniques (e.g., bicubic interpolation). Moreover, it accelerates SR reconstruction process by conducting computationally expensive convolutional operations ($N$ residual blocks) on $\times 8$ LR images/features \cite{Dong2016, LapSRN}. The deconvolved feature maps $\mathbf{F_{Re-F_N}}$ and $\mathbf{F_{Re-T}}$ are combined through an element-wise addition operation, and a convolution layer is applied to generate the super-resolved thermal image $I^{SR}$.

To drive the back-propagation to update the weights and biases of FL-MFRN model, we minimize the $L_1$ loss which computes the pixel-wise difference between the super-resolved thermal image $I^{SR}$ and the ground truth $I^{GT}$ as:
\begin{equation}
\mathcal{L}_{L_1}(P) = \sum_{p\in P}||I^{SR}(p) - I^{GT}(p)||_1,
\end{equation}
where $||\cdot||_1$ denotes the $L_1$ norm. We experimentally observed that our SR models achieve more accurate results using the $L_1$ loss, which is consistent with the previous finding that the $L_1$ loss is a better option than the conventional $L_2$ loss for image restoration and enhancement tasks \cite{EDSR}. 

\section{Experiments and comparisons}
\label{sec:Experiments and comparisons}

\subsection{Multispectral Dataset}
Using our multi-modal imaging system and the proposed pixel-wise alignment method, we generate 150 well-aligned visible and thermal image pairs in various indoor scenes. Note that the effective working distance of Intel Realsense RGB-D sensor ranges from 0.15 m to 8.0 m which is not enough for outdoor image capturing. To increase the diversity of the multispectral input, we also included 130 image pairs of traffic scenes from the publicly available MFNet multispectral dataset\cite{Ha2017MFNet}. Note effects of misalignment do not appear obvious for distant objects/backgrounds. In total, we obtain 280 pixel-wise aligned visible and thermal image pairs covering a wide range of contents (e.g., vehicle, machine, human, and building). 250 pairs are used as the training set and the rests are used as the testing set (20 pairs of our own captured images and 10 pairs from the MFNet dataset). Our captured multispectral dataset will be made publicly available in the future.
\begin{figure*}[ht]
	\footnotesize
	\centering
	\begin{minipage}[t]{0.16\linewidth}
		\includegraphics[width=1\linewidth]{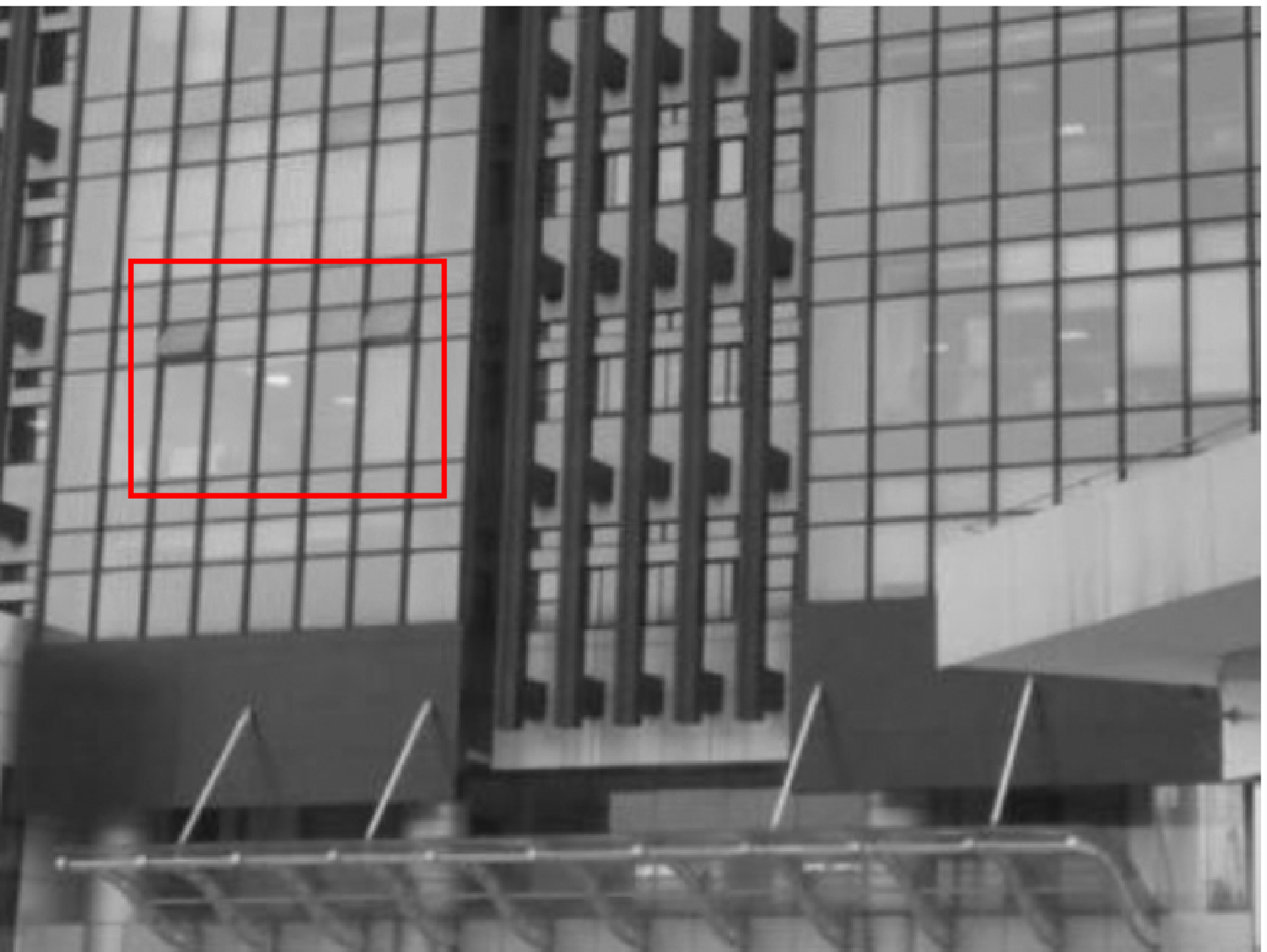}
	\end{minipage}
	\begin{minipage}[t]{0.16\linewidth}
		\includegraphics[width=1\linewidth]{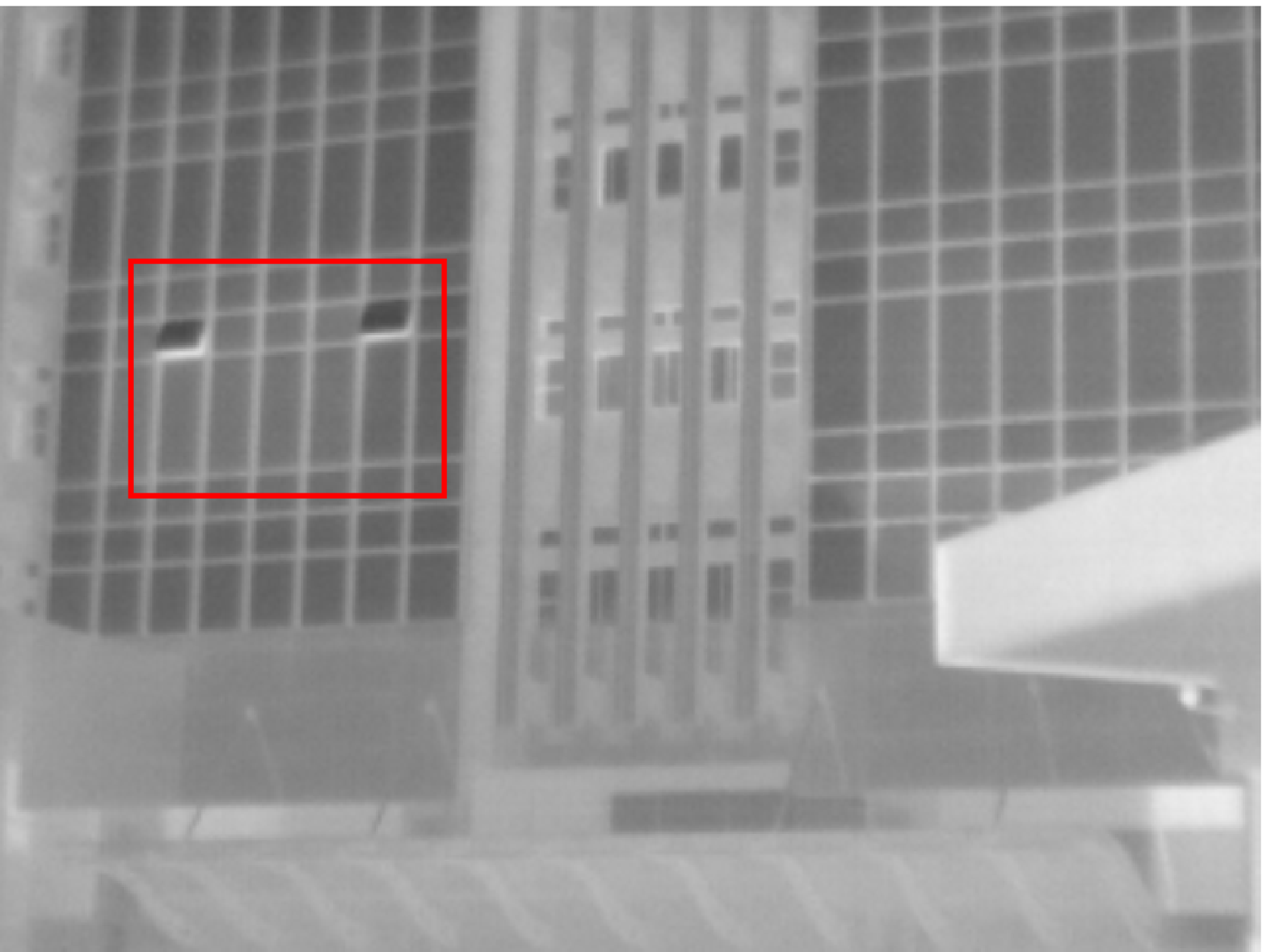}
	\end{minipage}
	\begin{minipage}[t]{0.16\linewidth}
		\includegraphics[width=1\linewidth]{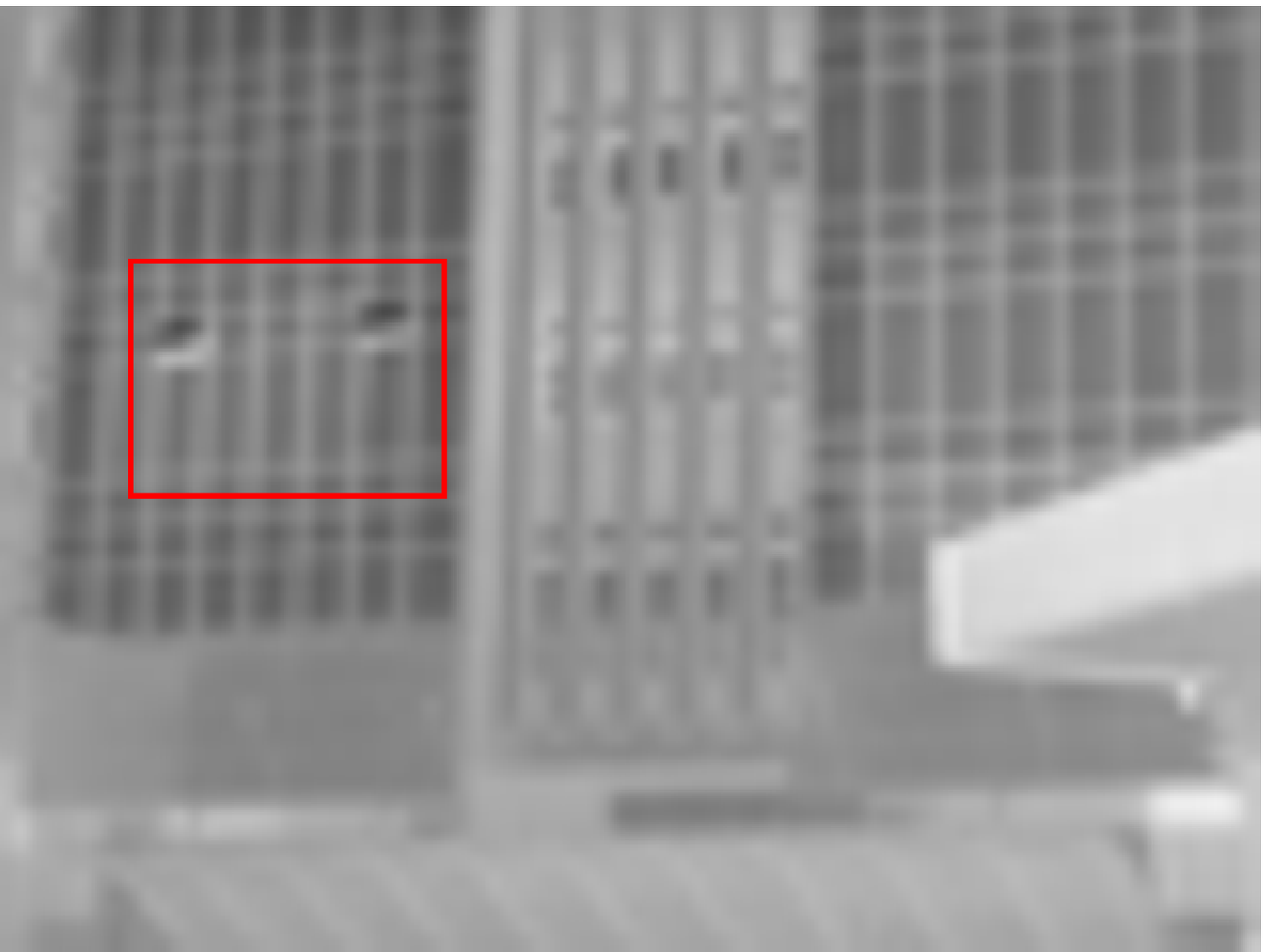}
	\end{minipage}
	\begin{minipage}[t]{0.16\linewidth}
		\includegraphics[width=1\linewidth]{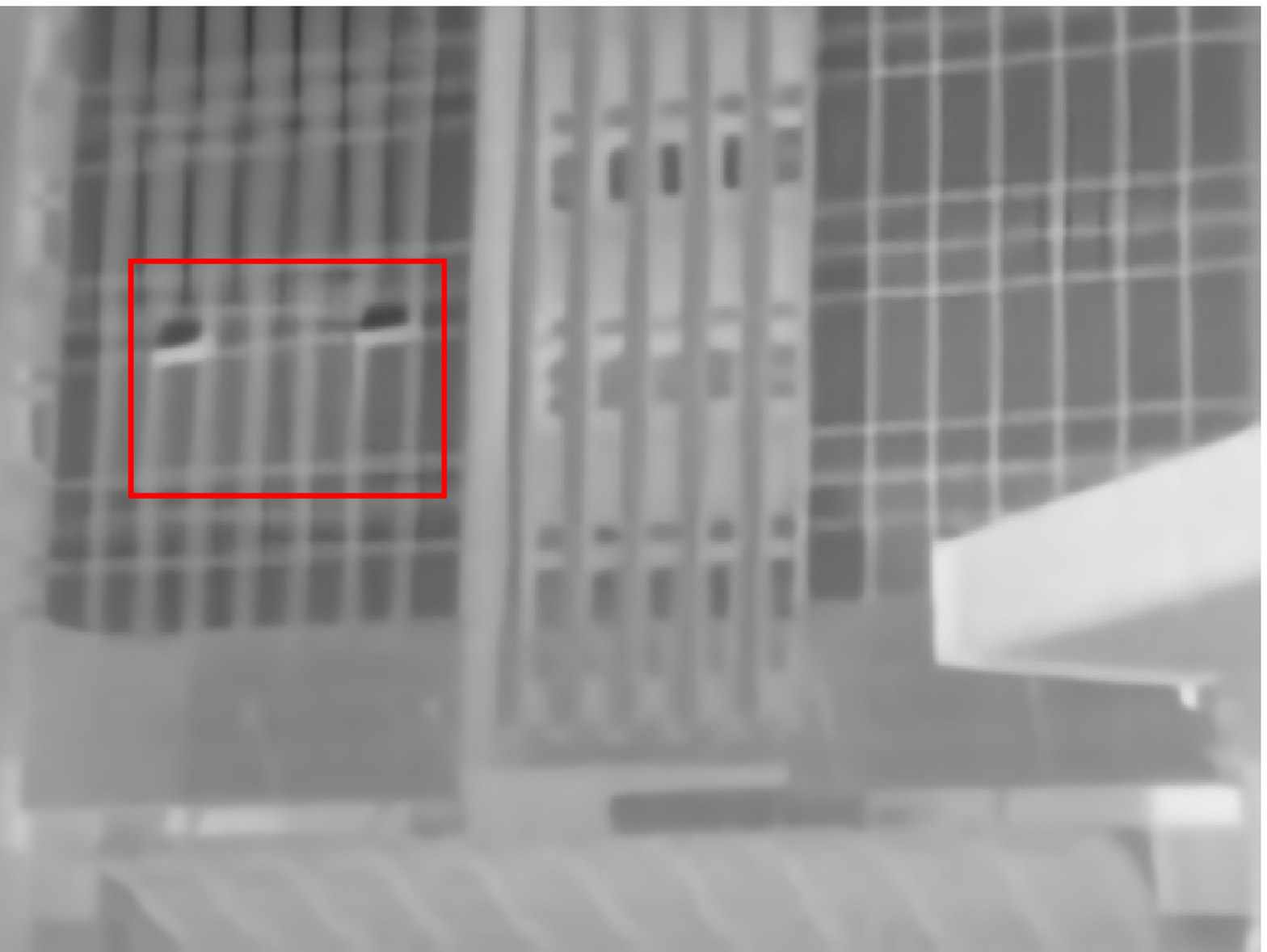}
	\end{minipage}
	\begin{minipage}[t]{0.16\linewidth}
		\includegraphics[width=1\linewidth]{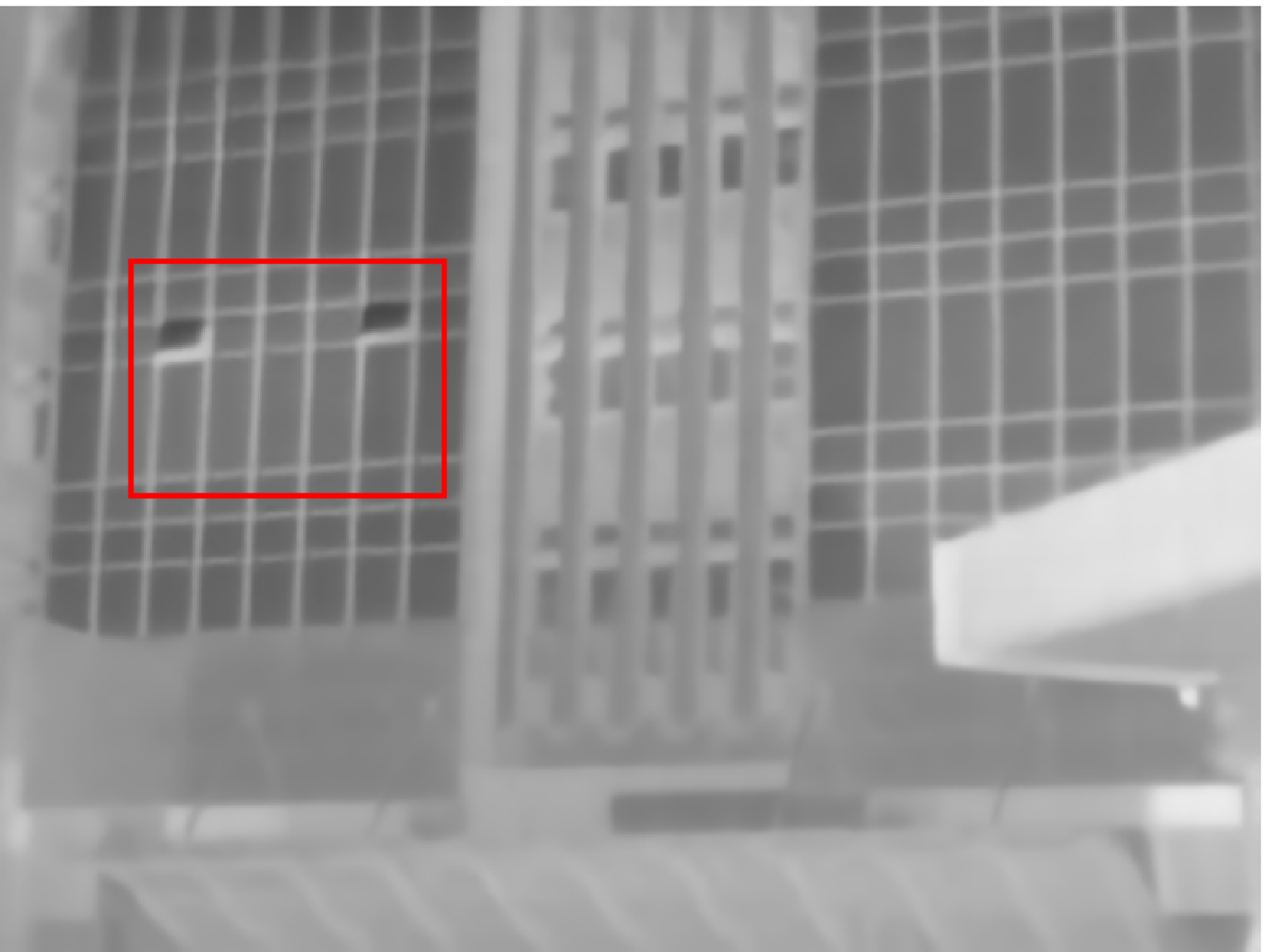}
	\end{minipage}\\
	 \vspace{1mm}	
	\begin{minipage}[t]{0.16\linewidth}
		\includegraphics[width=1\linewidth]{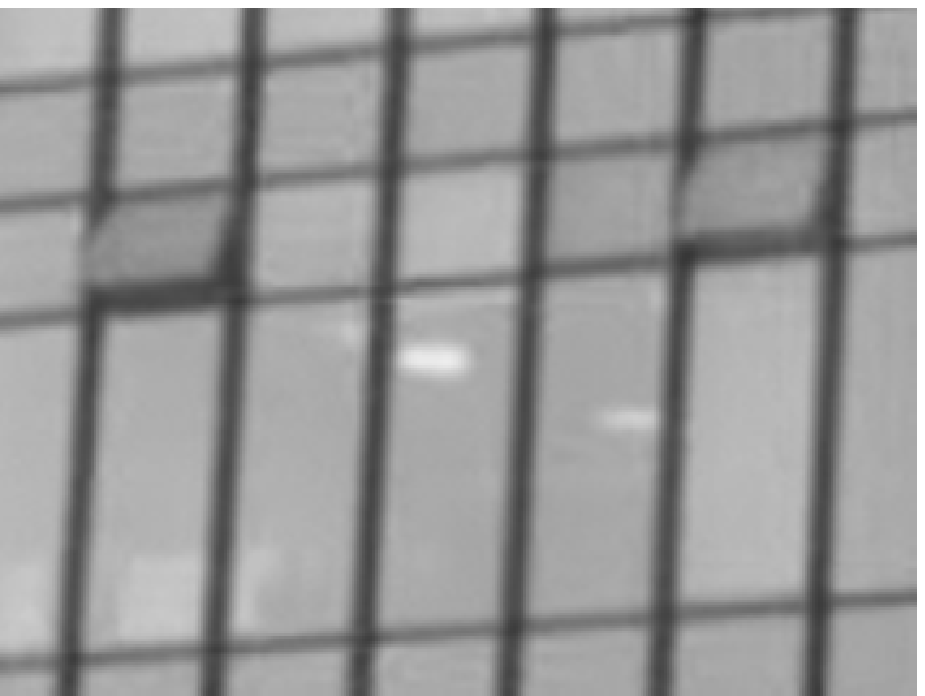}
		\centering{Y Channel}\\
		\centering{PSNR/SSIM}
	\end{minipage}
	\begin{minipage}[t]{0.16\linewidth}
		\includegraphics[width=1\linewidth]{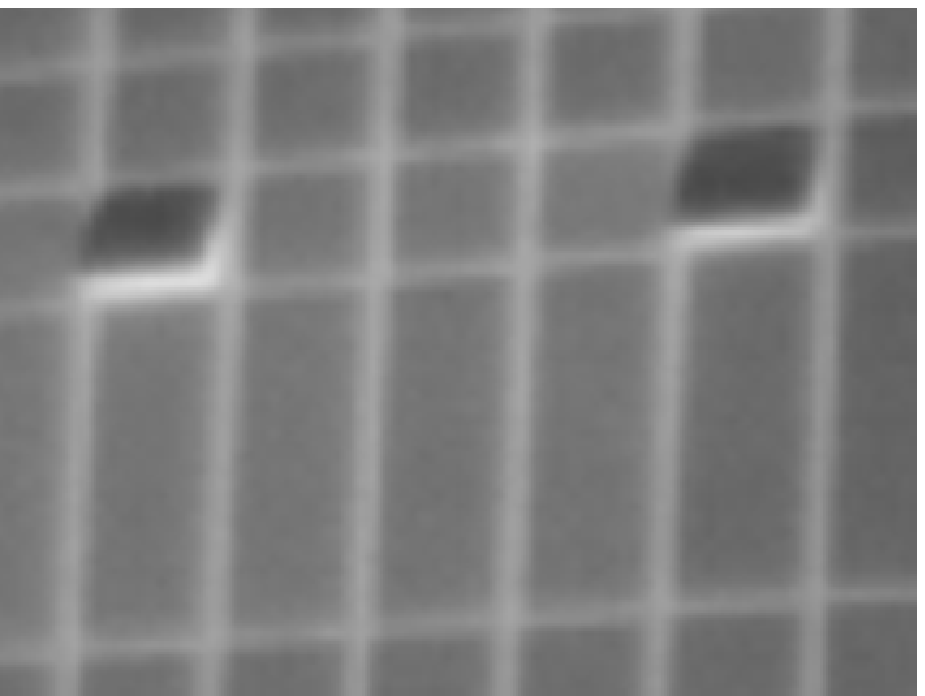}
		\centering{Thermal GT}\\
		\centering{-/-}
	\end{minipage}
	\begin{minipage}[t]{0.16\linewidth}
		\includegraphics[width=1\linewidth]{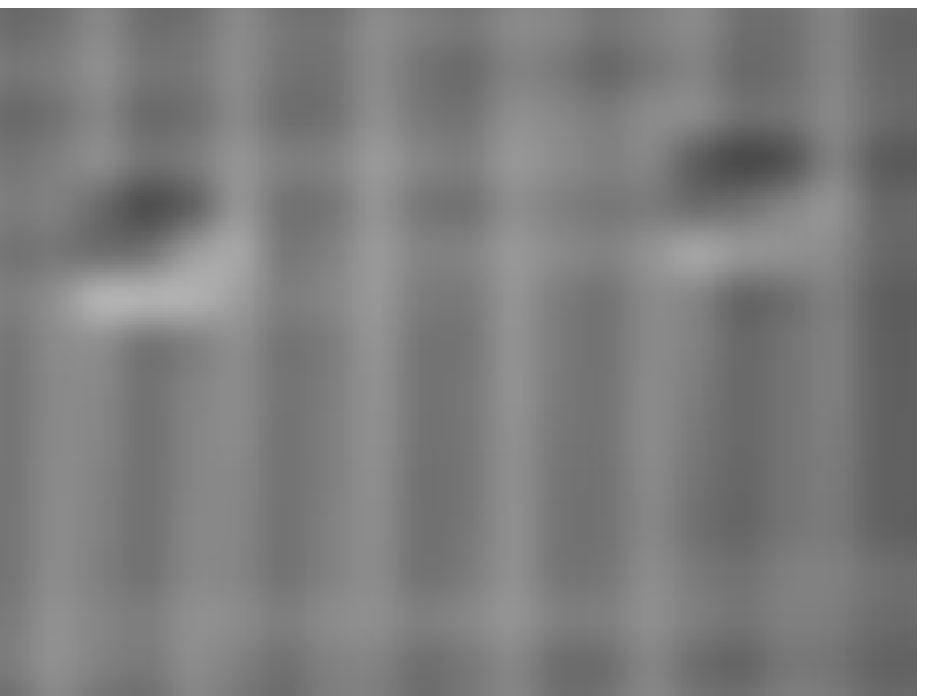}
		\centering{Bicubic}\\
		\centering{32.21/0.9319}
	\end{minipage}
	\begin{minipage}[t]{0.16\linewidth}
		\includegraphics[width=1\linewidth]{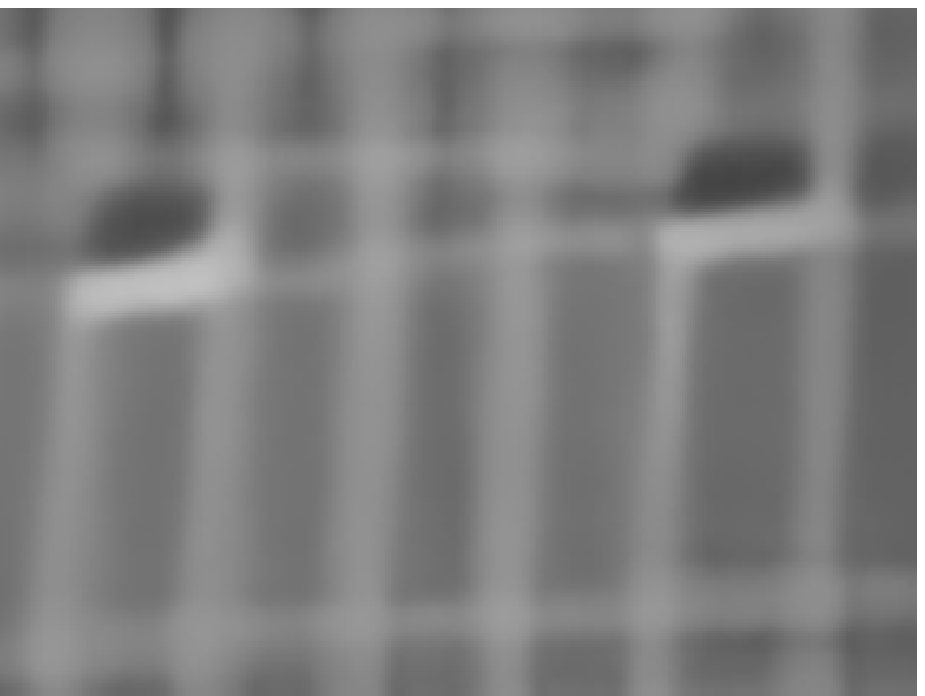}
		\centering{FL-MFRN-TT}\\
		\centering{34.18/0.9566}
	\end{minipage}
	\begin{minipage}[t]{0.16\linewidth}
		\includegraphics[width=1\linewidth]{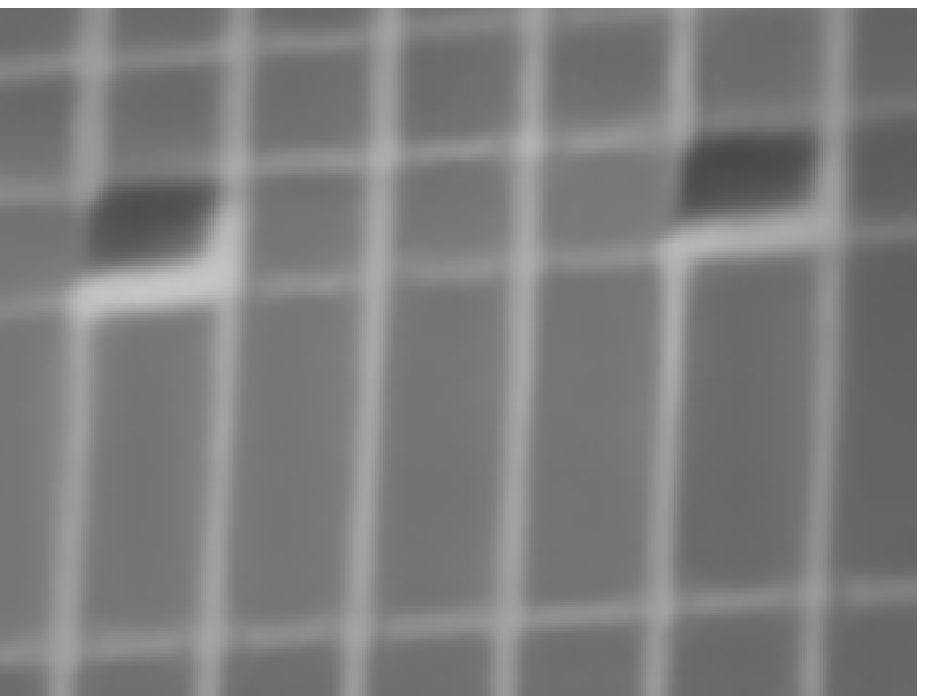}
		\centering{FL-MFRN-VT}\\
		\centering{35.33/0.9697}
	\end{minipage}
	\caption{Comparative SR results with/without deep fusion of multispectral images. Please zoom in to check details highlighted in the red region.}
	\label{fig:284}
\end{figure*}

\subsection{Implementation Details}
\label{subsec:implementation details}


To train our FL-MFRN model, we prepare a large number of training pairs, each of which consists a HR thermal image, a LR thermal image (down-sampling the HR thermal image via bicubic interpolation with a scale factor $8$), and a HR visible image. Three commonly-used data augmentation (e.g., rotation, flipping, and scaling) techniques are adopted to expand our training set.  After data augmentation, we randomly crop $96\times96$ image patches on the HR thermal and HR visible images and $12\times12$ image patches on the LR thermal images.


We apply Adam optimizer \cite{Adam} with a batch of 64 patches to minimize the $L_1$ loss. The initial learning rate is set to $1e-4$ and halves for every 20 epochs. We experimentally find 80 epochs are enough for achieving the best performance. All of our experiments are implemented on PyTorch platform with Cuda 9.1 and Cudnn 7.6.5. It takes approximately 16 hours to train our proposed FL-MFRN (containing 16 residual blocks) with scale factor $\times8$ on a single GPU of NVIDIA RTX 2080Ti (12 GB memory).

PSNR and SSIM  indexes are adopted to evaluate the SR performance quantitatively \cite{Wang2004}. Higher PSNR and SSIM values indicate more accurate SR results. For a fair comparison, we crop off boundary image pixels according to \cite{SRCNN2015PAMI}.

\begin{figure}[ht]
	\footnotesize
	\centering
	\begin{minipage}[t]{0.32\linewidth}
		\includegraphics[width=1\linewidth]{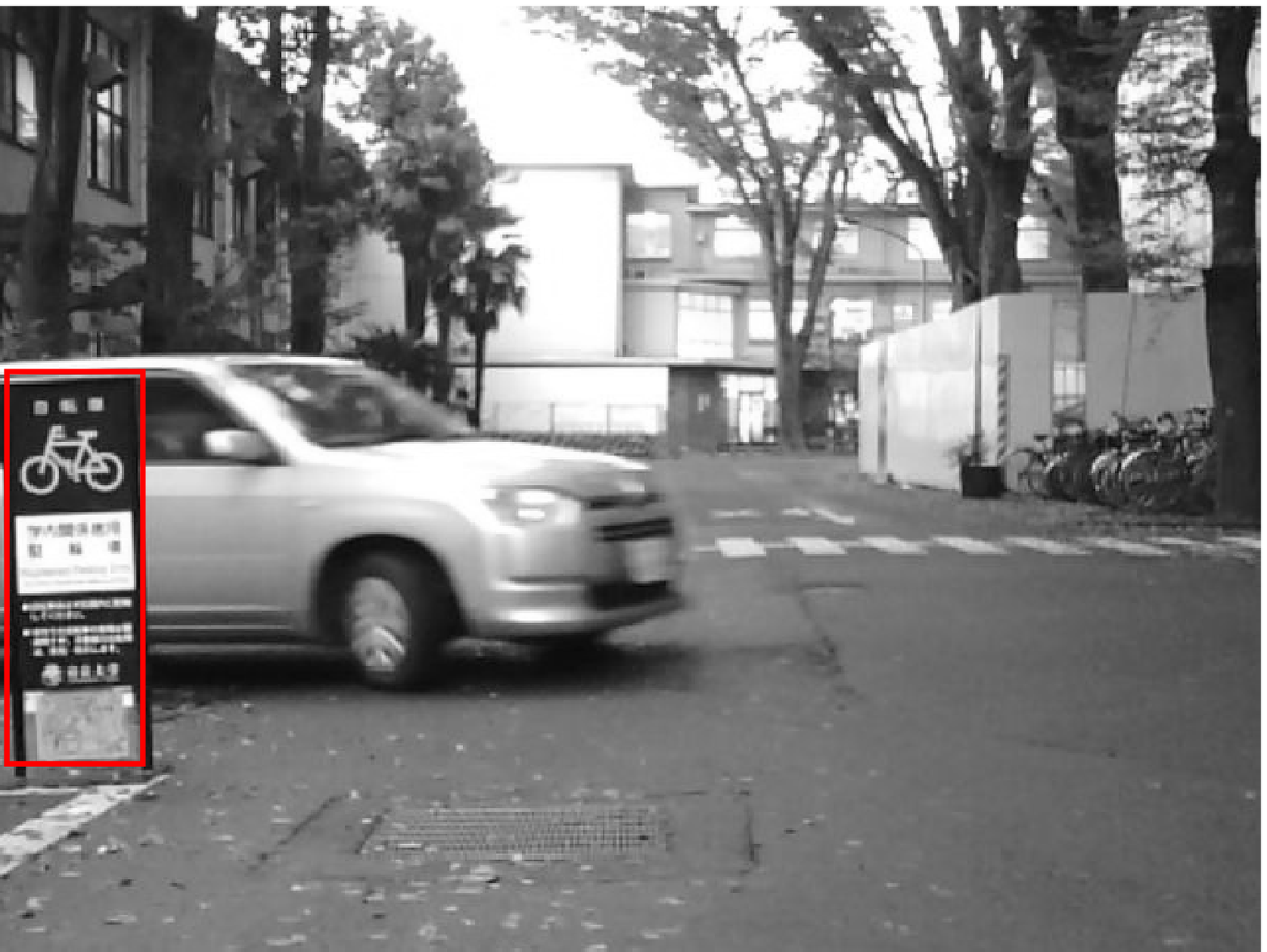}
		\centering{GT Y channel}\\
	\end{minipage}
	\begin{minipage}[t]{0.32\linewidth}
		\includegraphics[width=1\linewidth]{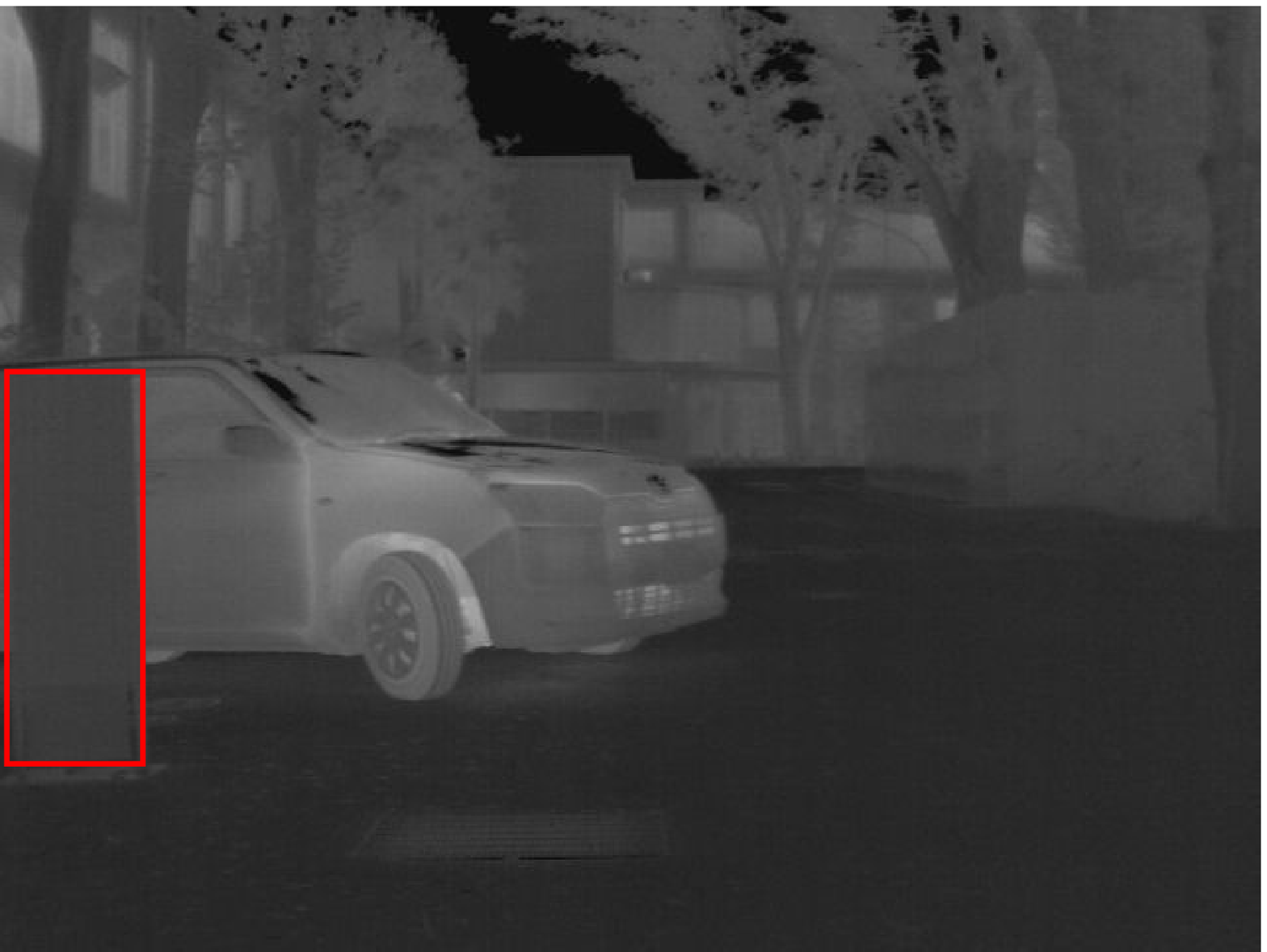}
		\centering{GT Thermal}\\
	\end{minipage}
	\begin{minipage}[t]{0.32\linewidth}
		\includegraphics[width=1\linewidth]{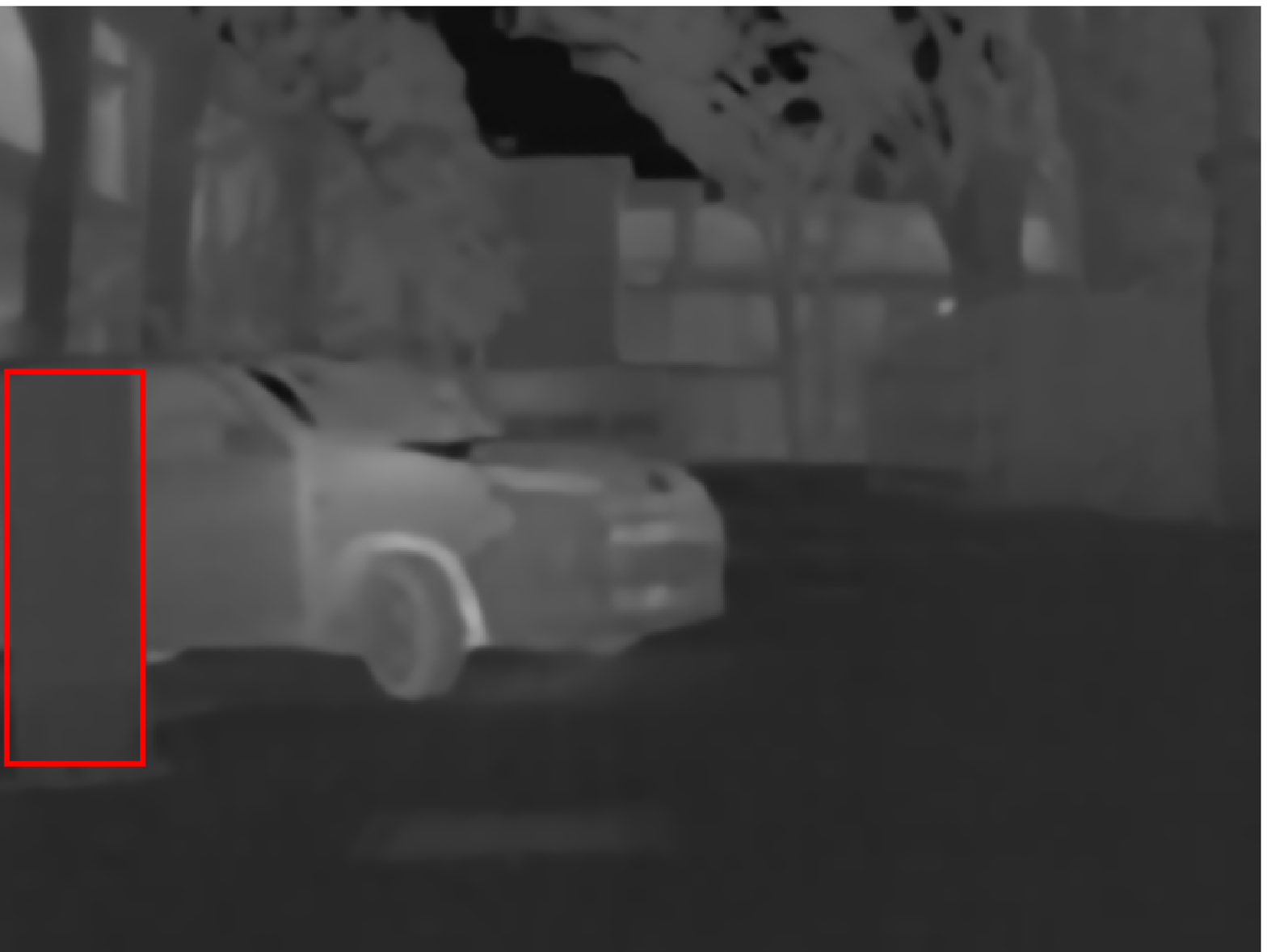}
		\centering{FL-MFRN}\\
	\end{minipage}	\\
    \caption{FL-MFRN model can accurately restore fine thermal details without adding irrelevant high-frequency signals extracted in visible images. Please zoom in to check details highlighted in the red region.}
	\label{fig:6D}
\end{figure}

%
\begin{figure*}[ht]
	\footnotesize
	\centering
	\begin{minipage}[t]{0.16\linewidth}
		\includegraphics[width=1\linewidth]{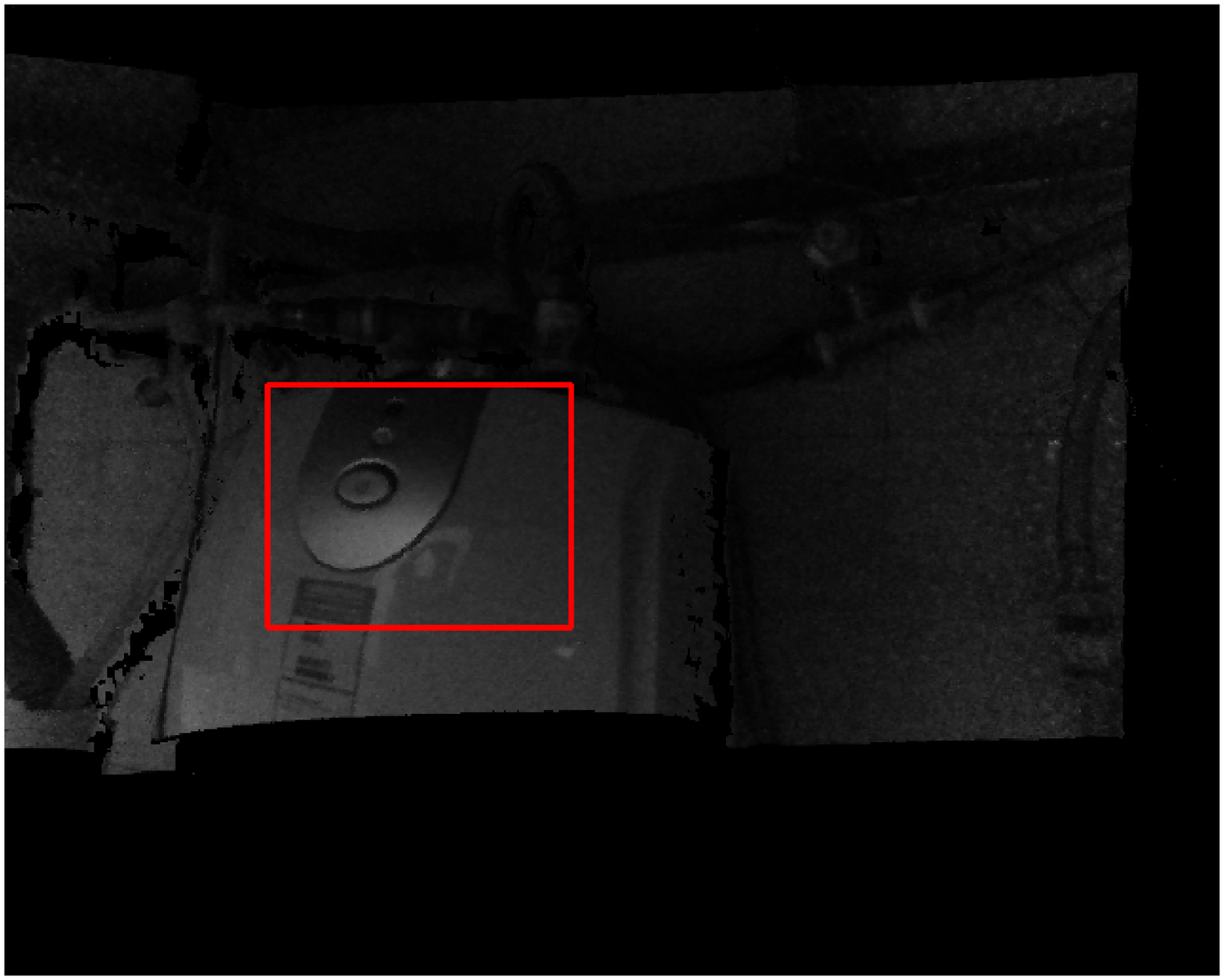}
	\end{minipage}
	\begin{minipage}[t]{0.16\linewidth}
		\includegraphics[width=1\linewidth]{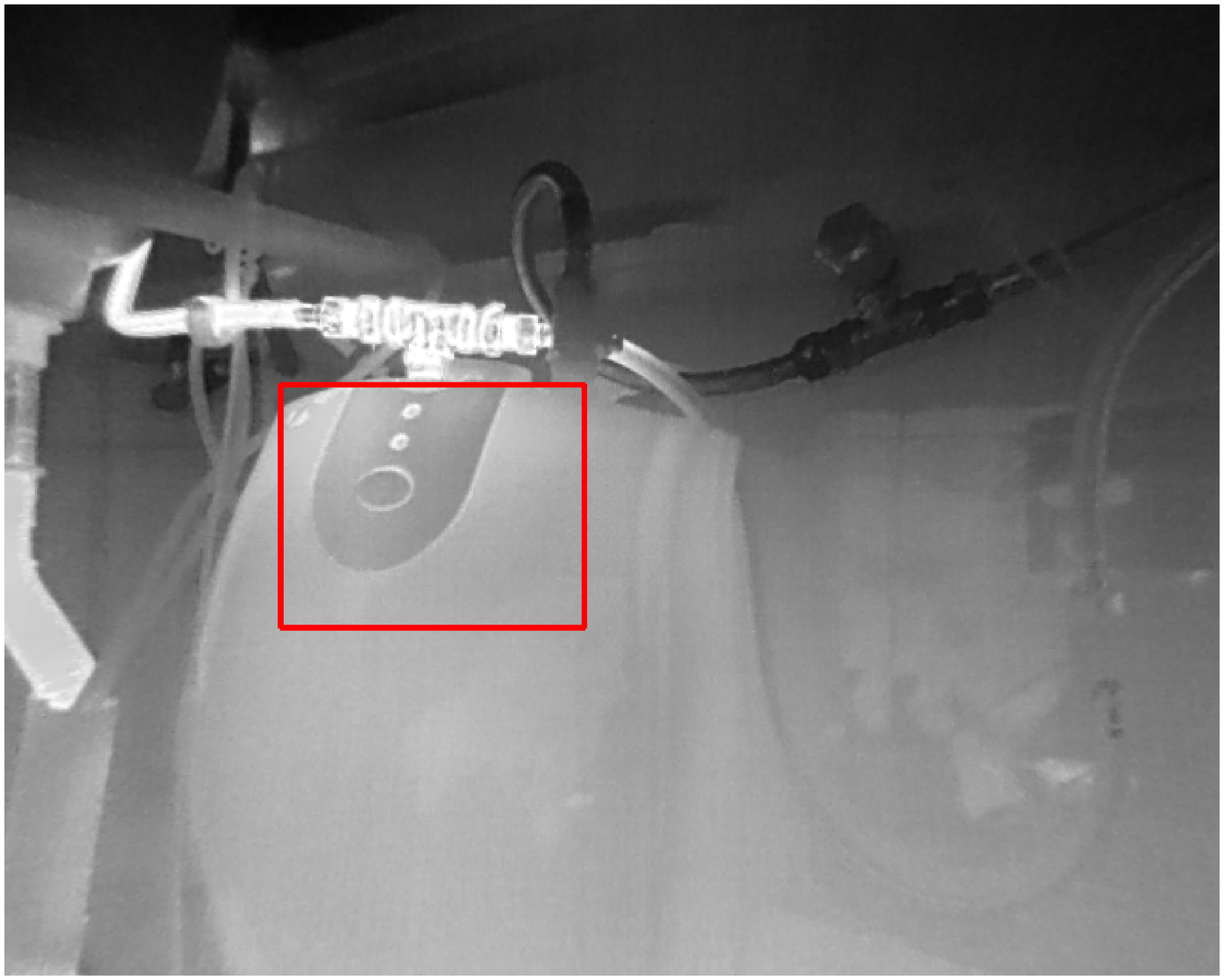}
	\end{minipage}
	\begin{minipage}[t]{0.16\linewidth}
		\includegraphics[width=1\linewidth]{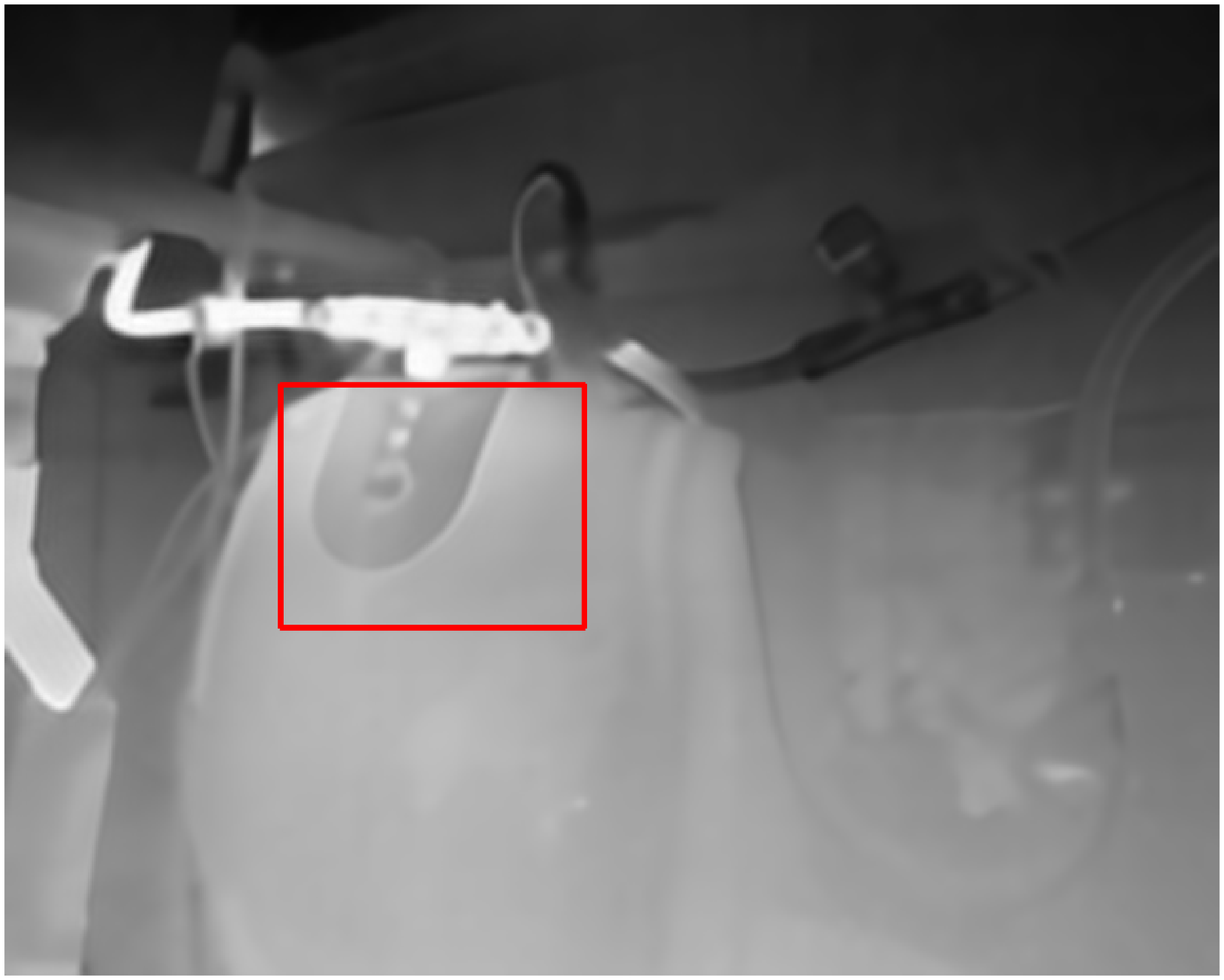}
	\end{minipage}	
	\begin{minipage}[t]{0.16\linewidth}
		\includegraphics[width=1\linewidth]{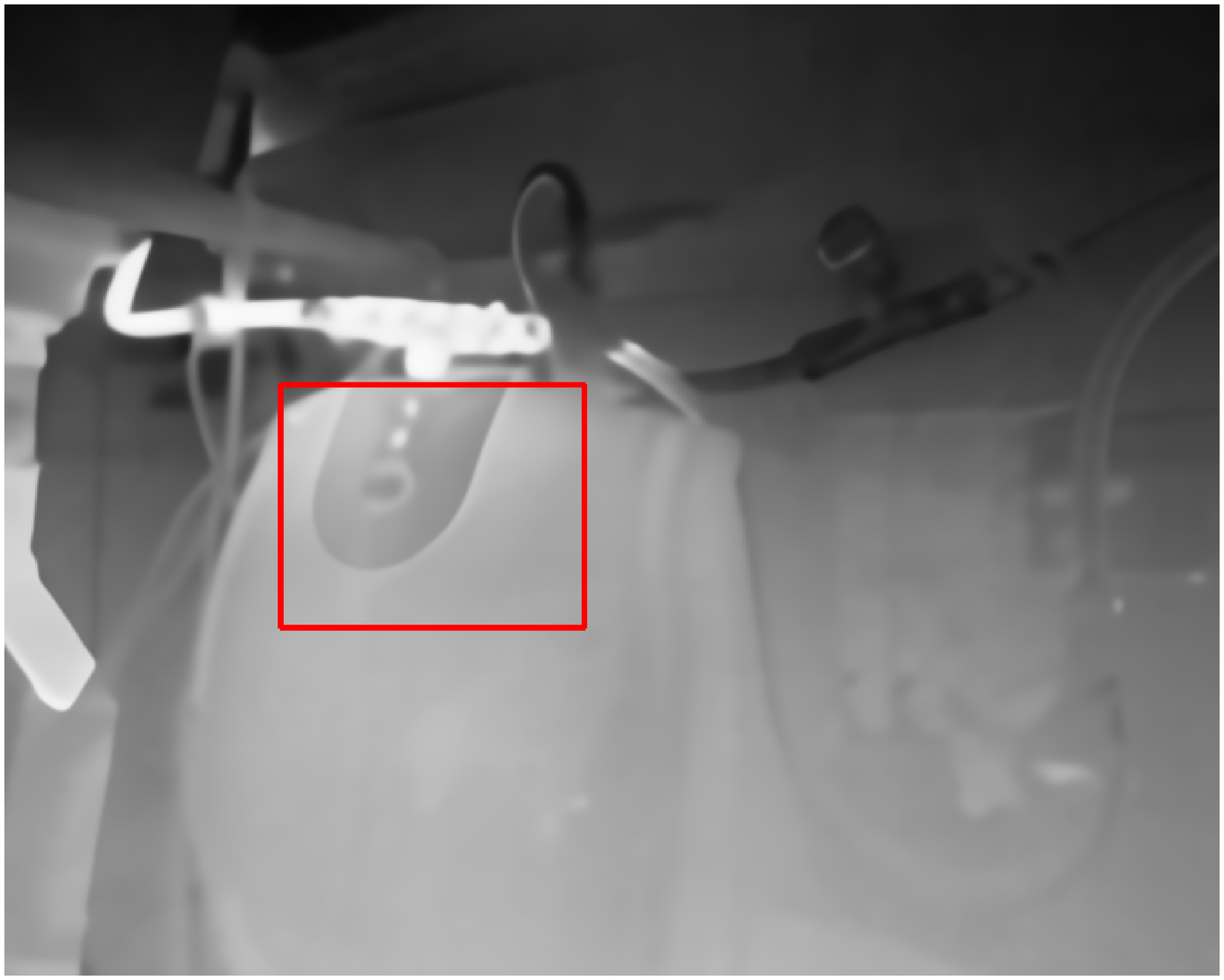}
	\end{minipage}
	\begin{minipage}[t]{0.16\linewidth}
		\includegraphics[width=1\linewidth]{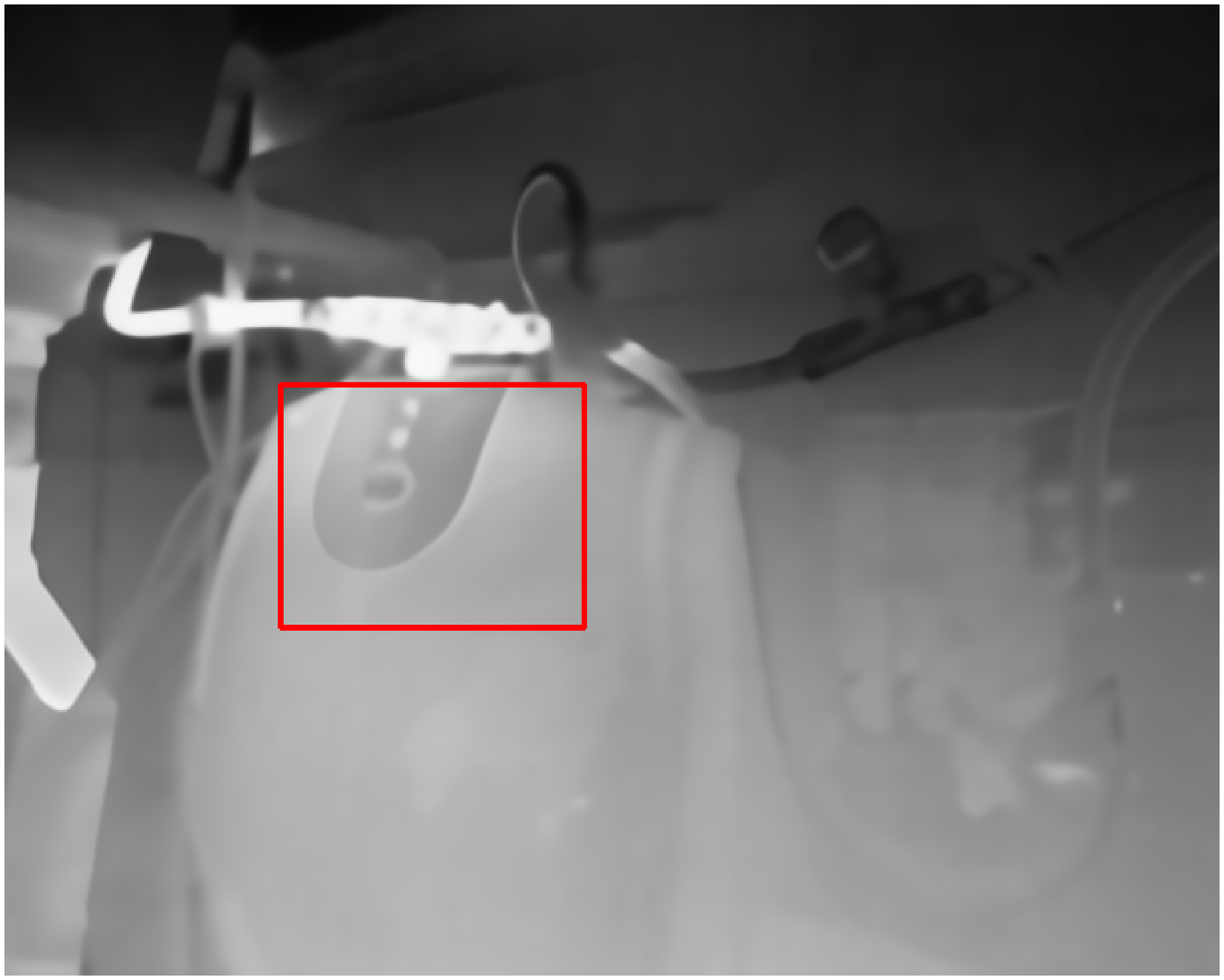}
	\end{minipage}
	\begin{minipage}[t]{0.16\linewidth}
		\includegraphics[width=1\linewidth]{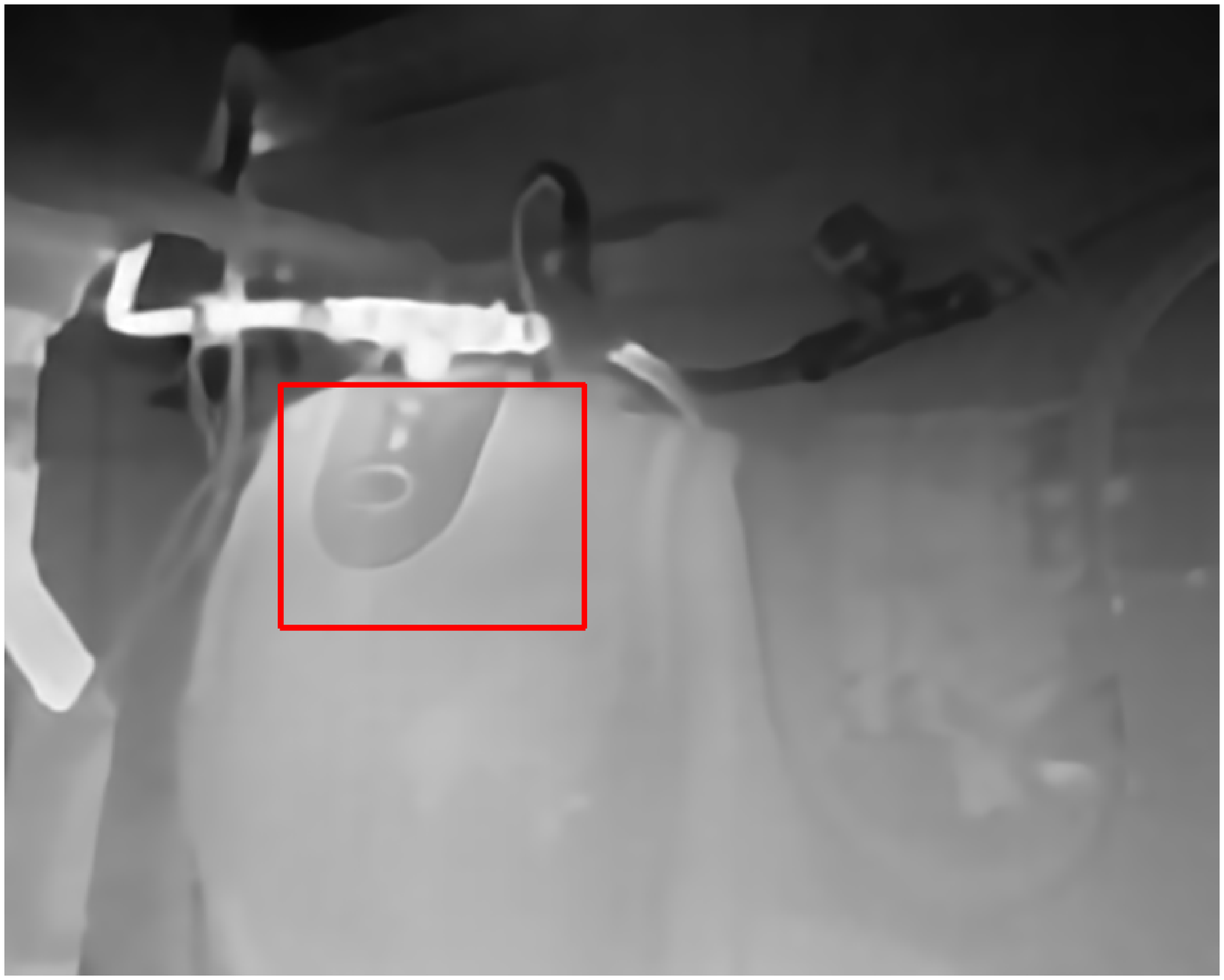}
	\end{minipage}\\
	 \vspace{1mm}	
	\begin{minipage}[t]{0.16\linewidth}
		\includegraphics[width=1\linewidth]{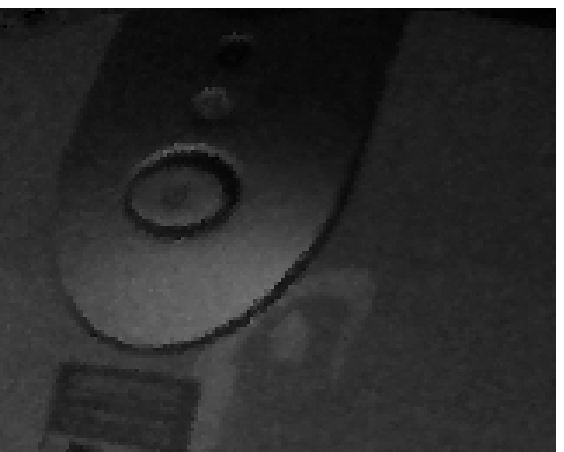}
		\centering{Y Channel}\\
		\centering{PSNR/SSIM}
	\end{minipage}
	\begin{minipage}[t]{0.16\linewidth}
		\includegraphics[width=1\linewidth]{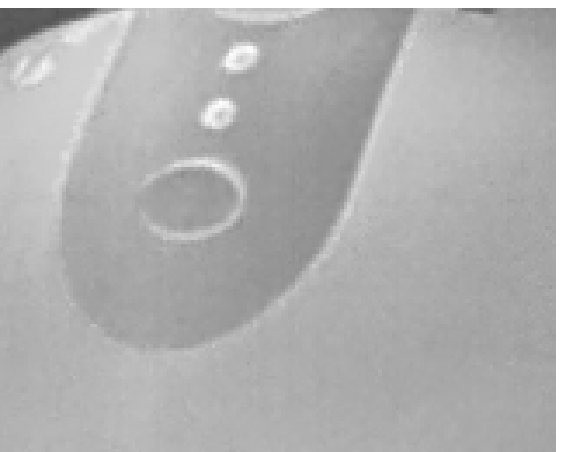}
		\centering{Thermal GT}\\
		\centering{-/-}
	\end{minipage}
	\begin{minipage}[t]{0.16\linewidth}
		\includegraphics[width=1\linewidth]{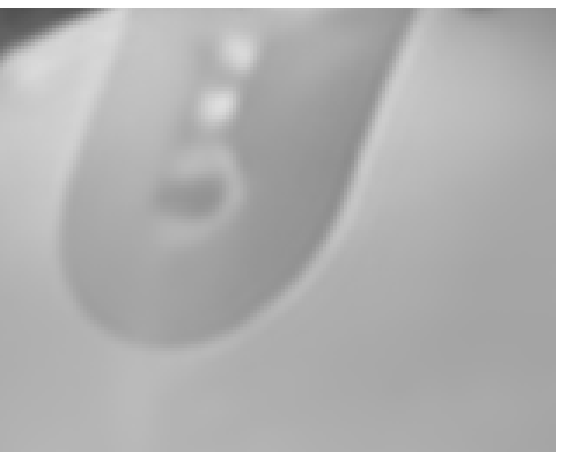}
		\centering{SAN \cite{dai2019second}}\\
		\centering{34.22/0.9094}
	\end{minipage}
	\begin{minipage}[t]{0.16\linewidth}
		\includegraphics[width=1\linewidth]{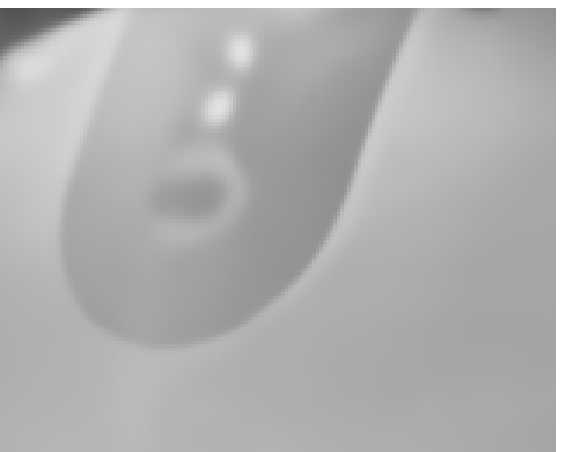}
		\centering{DRLN \cite{anwar2020densely}}\\
		\centering{34.37/0.9113}
	\end{minipage}
	\begin{minipage}[t]{0.16\linewidth}
		\includegraphics[width=1\linewidth]{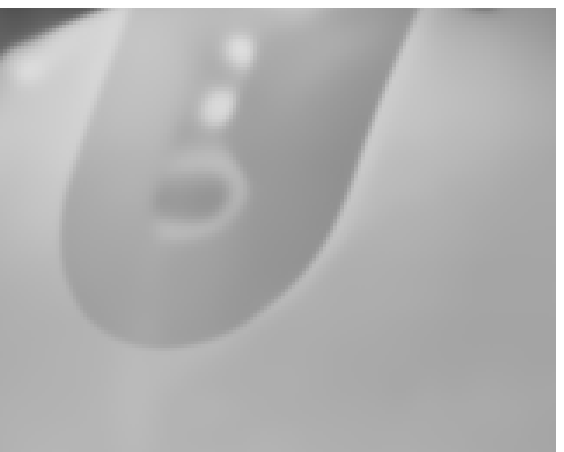}
		\centering{RCAN \cite{zhang2018image}}\\
		\centering{34.34/0.9106}
	\end{minipage}
	\begin{minipage}[t]{0.16\linewidth}
		\includegraphics[width=1\linewidth]{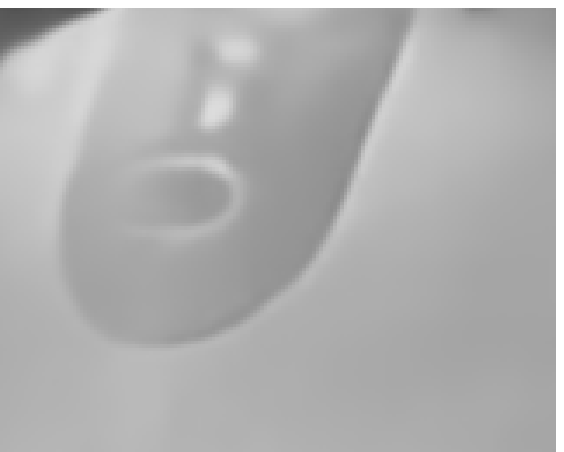}
		\centering{FL-MFRN}\\
		\centering{34.68/0.9238}
	\end{minipage}
    \caption{SR results using SAN \cite{dai2019second}, DRLN \cite{anwar2020densely}, RCAN\cite{zhang2018image} and our method. Please zoom in to check details highlighted in the red rectangles.}
	\label{fig:comparison1}
\end{figure*}

\subsection{Effectiveness of Multispectral Fusion}
\label{Effectiveness}
For this purpose, we trained two different versions of FL-MFRN models. The first model FL-MFRN-VT is based on a HR visible and a LR thermal images and the second model FL-MFRN-TT is based on the thermal channel only (feeding the visible feature extraction stream using a bicubic interpolation up-sampling thermal image). For a fair comparison, we adopt the concatenation fusion scheme $f^{\text{sum}}$ to combine visible and thermal low-level features for both models. We also show some qualitative SR results of FL-MFRN-VT and FL-MFRN-TT models in Fig.~\ref{fig:284}. Texture/edge features in the visible image can assist to accurately recover important thermal details (e.g., tree branches and leaves) which are difficult/impossible to restored based on LR thermal images alone. On 30 pairs of testing images, FL-MFRN-VT model achieves significantly higher SR accuracy than FL-MFRN-TT model (PSNR: 33.33dB vs. 32.79dB and SSIM: 0.8599 vs. 0.8436).

In Fig.~\ref{fig:6D}, we show another example of SR result when some salient objects are only presented in the visible channel but not in the thermal channel. It is observed that FL-MFRN model, which considers both temperature profile in the thermal channel and texture/edge features in the visible channel, can accurately restore fine thermal details without adding irrelevant high-frequency signals extracted in visible images.  The experimental results qualitatively and quantitatively show the deep fusion of multispectral data can achieve better SR results in terms of higher accuracy and better visual perception.

\subsection{Comparisons with State-of-the-arts}
\label{comparisons with sota}

We perform quantitative and qualitative experiments to compare our proposed FL-MFRN with state-of-the-art CNN-based SR approaches including SAN \cite{dai2019second}, DRLN \cite{anwar2020densely}, and RCAN \cite{zhang2018image}. Source codes or pre-trained models of these methods are publicly available. To achieve optimal SR performance on thermal images, we fine-tuned these CNN-based SR models using the captured multispectral dataset. Moreover, we re-implemented a visible-channel guided thermal image SR solution (CGSR \cite{chen2016color}). 

Tab.~\ref{sota} show the quantitative evaluation results (PSNR, SSIM, running time, and size) of our method and the other SR approaches. It is noted that FL-MFRN can achieve more accurate SR results, producing higher PSNR values (CGSR \cite{chen2016color}: 5.19
dB$\uparrow$,  SAN \cite{dai2019second}: 0.67dB$\uparrow$,  DRLN \cite{anwar2020densely}: 0.41dB$\uparrow$, RCAN\cite{zhang2018image}: 0.44dB$\uparrow$). Another advantage of our proposed FL-MFRN model is that it can achieve better SR accuracy using significantly less running time and fewer parameters. We calculate the averaged running time of different SR methods to process 100 input images on a PC equipped with 2.1 GHz Intel Xeon E5-2620V4 CPU (64 GB RAM) and NVIDIA RTX 2080Ti (12 GB memory) GPU. Note FL-MFRN model processes pairs of $640\times512$ visible and $80\times64$ thermal images, while other CNN-based SR models only process LR thermal images.

Some comparative results are provided in Fig.~\ref{fig:comparison1}. It is visually observed that our proposed FL-MFRN model achieves more accurate thermal image SR results. As highlighted in red regions, FL-MFRN model is capable of recovering high-fidelity thermal details (e.g., window frames, facade boundaries) by considering texture/edge features in the visible channel, which are difficult/impossible to restored based on $\times8$ thermal images alone. 

\begin{table}[ht]
	\footnotesize
	\centering
		\caption{Quantitative evaluation results of our method and the state-of-the-art SR approaches \cite{chen2016color, dai2019second, anwar2020densely, zhang2018image}.}
	\begin{tabular}{|c||c|c|c|c|}
		\hline
		Methods & PSNR (dB)   & SSIM   & Time (ms)   & Size (MB) \\ 
		\hline\hline
		BICUBIC & 28.13  & 0.7957      & /   &/ \\ 
		CGSR \cite{chen2016color} & 28.14  & 0.7919      & /   &/ \\ 
		 SAN \cite{dai2019second} & 32.66	& 0.8526 &	450 &	64.2 \\ 
		DRLN \cite{anwar2020densely}  & 32.92	& 0.856	&725	&139.1  \\ 
		RCAN\cite{zhang2018image}    & 32.89  &	0.8533  &	890  &	63.2 \\
		\hline
		\textbf{FL-MFRN}  & \textbf{33.33} &	\textbf{0.8599}	& \textbf{32}	& \textbf{6.5} \\

		\hline 
	\end{tabular}

		\label{sota}
\end{table}

\section{Conclusion}
\label{sec:Conclusion}

In this paper, we present a new framework to virtually generate pixel-wise aligned multispectral images and then perform deep fusion of multispectral images (visible and thermal) for high-accuracy image SR. Different from most existing SR methods which are trained and tested on a single spectral channel, our proposed method can boost the accuracy of SR in an expensive channel by utilizing information in a low-cost channel. Moreover, it can accurately restore fine thermal details without adding irrelevant high-frequency signals extracted in visible images. Experimental results on a newly constructed pixel-wise aligned multispectral dataset demonstrate the effectiveness of multispectral data fusion for SR task, achieving better SR results in terms of both higher accuracy and less running time. In the future, we plan to implement the proposed method to improve the quality of infrared images for other high-level thermal imaging applications including medical diagnosis, night vision, target detection, and object recognition.
 
%




%
%
%
%



\bibliographystyle{IEEEtran}
\bibliography{egbib}

\end{document}